\documentclass[runningheads]{llncs}
\usepackage[T1]{fontenc}
\usepackage{booktabs}
\usepackage{tabularx}
\usepackage{multirow}
\usepackage{multirow}
\usepackage{amsmath, amssymb}
\usepackage{cite}
\usepackage{hhline}
\usepackage{soul}
\usepackage{float}
\usepackage{booktabs}
\soulregister{\cite}7 
\soulregister{\citep}7 
\soulregister{\citet}7 
\soulregister{\ref}7 
\soulregister{\pageref}7 
\soulregister{\say}7 

\usepackage[colorlinks=true, citecolor=blue]{hyperref}

\usepackage{colortbl}
\usepackage[table,xcdraw]{xcolor}
\usepackage{soul}  
\usepackage{caption}

\usepackage{colortbl}
\definecolor{naturelight}{rgb}{0.6, 0.8, 0.4}
\definecolor{naturemedium}{rgb}{0.2, 0.5, 0.2}
\definecolor{naturedark}{rgb}{0.1, 0.3, 0.1}
\usepackage{dirtytalk}

\usepackage{graphicx}
\usepackage{colortbl}
\definecolor{naturelight}{rgb}{0.8, 1.0, 0.8} 
\definecolor{naturemedium}{rgb}{0.6, 0.8, 0.6} 
\definecolor{naturedark}{rgb}{0.4, 0.6, 0.4} 
\usepackage{booktabs}
\usepackage{adjustbox}
\usepackage{amssymb}
\usepackage{todonotes}
\usepackage{multirow}

\usepackage{hyperref}
\usepackage{marvosym}
\usepackage{float} 
\usepackage{marvosym} 
\usepackage[T1]{fontenc}

\definecolor{cgreen}{rgb}{0,0.7,0.8}
\definecolor{cred}{rgb}{0.968,0.545,0.321}
\definecolor{brightpink}{rgb}{1.0, 0.0, 0.5}

\renewcommand{\thefootnote}{\fnsymbol{footnote}} 
\usepackage{graphicx,verbatim}

\begin{document}

\title{Enhancing Abnormality Grounding for Vision Language Models with Knowledge Descriptions}

\author{Jun Li\inst{1,2} \and Che Liu\inst{4} \and  Wenjia Bai\inst{4} \and Rossella Arcucci\inst{4} \and \\
 Cosmin I. Bercea*\inst{1,3}\textsuperscript{(\Letter)}  
 \and Julia A.Schnabel*\inst{1,2,3,5}\textsuperscript{(\Letter)}} 

\authorrunning{Jun Li et al.}
\renewcommand{\thefootnote}{\fnsymbol{footnote}} 
\footnotetext[1]{Shared senior authors.}

\institute{Technical University of Munich, Germany  \and
Munich Center for Machine Learning, Germany \and
Helmholtz AI and Helmholtz Munich, Germany \and
Imperial College London, UK \and
King’s College London, UK \\
\email{\{june.li,cosmin.bercea,julia.schnabel\}@tum.de}
}

\maketitle     

\begin{abstract}
Visual Language Models (VLMs) have demonstrated impressive capabilities in visual grounding tasks. However, their effectiveness in the medical domain, particularly for abnormality detection and localization within medical images, remains underexplored. A major challenge is the complex and abstract nature of medical terminology, which makes it difficult to directly associate pathological anomaly terms with their corresponding visual features.
In this work, we introduce a novel approach to enhance VLM performance in medical abnormality detection and localization by leveraging decomposed medical knowledge. Instead of directly prompting models to recognize specific abnormalities, we focus on breaking down medical concepts into fundamental attributes and common visual patterns. This strategy promotes a stronger alignment between textual descriptions and visual features, improving both the recognition and localization of abnormalities in medical images.We evaluate our method on the 0.23B Florence-2 base model and demonstrate that it achieves comparable performance in abnormality grounding to significantly larger 7B LLaVA-based medical VLMs, despite being trained on only 1.5\% of the data used for such models. Experimental results also demonstrate the effectiveness of our approach in both known and previously unseen abnormalities, suggesting its strong generalization capabilities. 
The code and model are available \href{https://lijunrio.github.io/AG-KD/}{here}\footnote{Link: https://lijunrio.github.io/AG-KD/}.

\end{abstract}

\begin{figure*}[h]
\centerline{\includegraphics[width=0.9\textwidth]{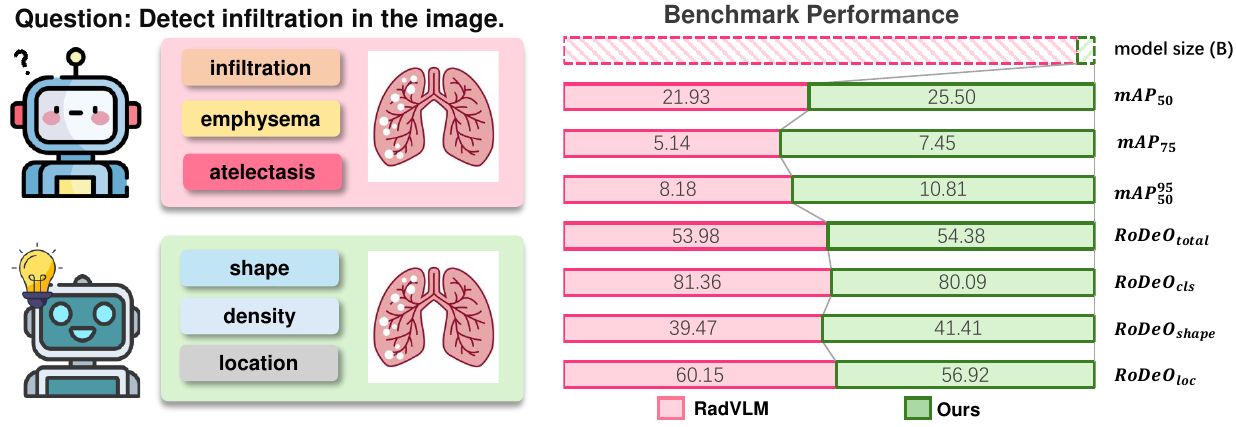}}
\caption{Overview of our approach. We train a 0.23B model on just 16,087 samples (1.5\% of the data) and achieve similar or better results than the 7B RadVLM, pre-trained on 1 million samples, by using text descriptions that highlight key visual features of abnormalities.}
\label{teaser}
\end{figure*}

\section{Introduction}
Vision-Language Models (VLMs)~\cite{openai2024chatgpt,xiao2024florence,bai2023qwen,lu2024deepseek} have achieved remarkable success in a variety of visual understanding tasks, such as image captioning~\cite{stefanini2022show}, visual question answering~\cite{lin2023medical}, and visual grounding~\cite{xiao2024towards}. By jointly modeling both visual and textual representations, these models excel at associating image content with natural language descriptions, enabling them to be highly adaptable across diverse domains. Recent advancements have extended VLMs to the medical imaging domain, where models such as RadVLM \cite{deperrois2025radvlm}, MAIRA-2 \cite{bannur2024maira}, and ChexAgent \cite{chen2024chexagent} have demonstrated significant potential in tasks like radiology report generation, question answering, and abnormality grounding. These models use large-scale paired X-rays image-text datasets \cite{johnson2019mimic,wu2021chest,bustos2020padchest,castro2024padchest}, allowing them to perform different medical tasks, thus providing valuable support to radiologists in diagnosis.

Despite these advancements, abnormality grounding remains a critical yet underexplored task in medical image analysis. Unlike report generation\cite{li2024ultrasound, tanida2023interactive,li2022self} and question answering \cite{singhal2025toward,li2025language,zhang2023pmc}, which has been extensively studied within the context of VLMs, abnormality grounding requires not only the ability to understand textual queries but also the precise localization of corresponding abnormalities within images. Existing medical VLMs, though powerful, are typically large-scale models that require substantial computational resources and extensive pretraining on diverse datasets. While these models demonstrate strong general performance, their general-purpose design may limit their ability to effectively address specialized tasks such as abnormality grounding on medical images. Moreover, a significant challenge in abnormality grounding stems from the complexity of medical terminology and its weak alignment with visual features. In general visual grounding tasks, objects like \say{cat} or \say{dog} possess well-defined and easily recognizable features, facilitating the formation of direct visual-language associations. However, medical abnormalities present more nuanced challenges. Terms like \say{lung opacity} or \say{interstitial lung disease} refer to combinations of textural, morphological, and contextual features, none of which map to a single, well-defined visual counterpart. Instead, these terms describe subtle, heterogeneous manifestations that can vary based on clinical context and imaging modalities.

To address these challenges, we introduce a novel approach for abnormality grounding by incorporating decomposed knowledge descriptions tied to visual features as shown in Figure \ref{teaser}. Specifically, we leverage textual descriptions capturing key visual attributes of medical abnormalities, including shape, density, and location—essential for accurate abnormality interpretation in medical images. By explicitly encoding this domain-specific knowledge, we enhance the model’s ability to associate complex medical terms with their corresponding visual features. Our method achieves superior performance compared to state-of-the-art approaches\cite{deperrois2025radvlm, bannur2024maira}, despite utilizing a much smaller model framework with fewer parameters\cite{xiao2024florence}. Additionally, our approach demonstrates that knowledge-enhanced prompts can considerably boost performance in zero-shot settings, even for previously unseen abnormalities. Our key contributions are as follows:
\begin{itemize}
    \item We propose a knowledge-enhanced approach for abnormality grounding in VLM training, by using fine-grained, attribute-based textual descriptions to improve visual grounding performance.
    \item We show that a small-scale VLM (0.23B parameters) can match the abnormality grounding performance of large-scale models (7B parameters) using only 1.5\% of the training data.
    \item We show that our approach considerably improves zero-shot generalization, enabling the model to better detect unseen abnormalities, which is essential in low-data scenarios.
\end{itemize}

\begin{figure*}[!t]
\centerline{\includegraphics[width=\textwidth]{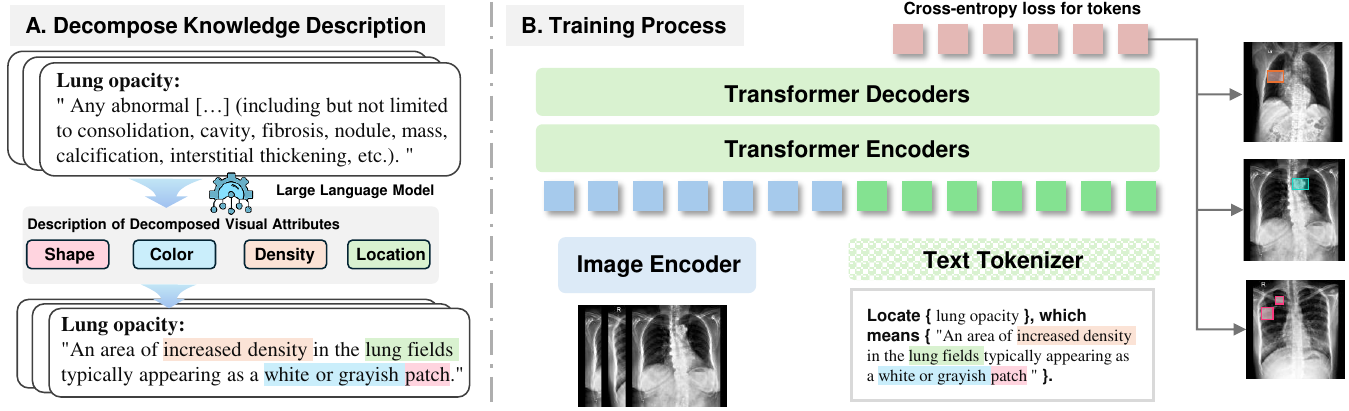}}
\caption{ Overview of our method. (A) shows the pipeline for obtaining decomposed knowledge descriptions, (B) presents the model architecture and training process for the abnormality grounding task.}
\label{method}
\end{figure*}

\section{Methods}
\subsubsection{Problem Setup.}

Unlike traditional object detection methods~\cite{jiang2022review} that regress bounding box coordinates, VLMs frame detection as an autoregressive sequence generation task. Given an image $I \in \mathbb{R}^{H \times W}$, where $H$ and $W$ denote its dimensions, and a text prompt $T$ describing the target object, the VLM generates a set of bounding boxes $L = \{ L_1, L_2, \dots, L_M \}$. Each bounding box $L_i = \{\ell_{x_0}^i, \ell_{y_0}^i, \ell_{x_1}^i, \ell_{y_1}^i \}$ consists of four discrete tokens representing its top-left and bottom-right corners. Coordinates are normalized to $[0, 1000]$ as  
$\ell_x = \frac{x_{\text{pixel}}}{W} \cdot 1000$, $\ell_y = \frac{y_{\text{pixel}}}{H} \cdot 1000$  
and quantized into a fixed vocabulary, as described in \cite{xiao2024florence}. The bounding box count $M$ equals the number of detected instances. This formulation enables detection to be naturally integrated into language modeling.

\subsubsection{Decomposed Knowledge Description}  
We improve VLM performance in detecting and localizing medical abnormalities by breaking down complex medical concepts into key visual attributes, such as shape, location, and density, as shown in Figure \ref{method}A. We first retrieve the medical definitions of each abnormality \cite{nguyen2022vindr}, which often lack explicit references to their visual manifestations in medical images. For example, the medical definition of \textit{lung opacity} is:  \textit{``Any abnormal focal or generalized opacity or opacities in lung fields (including but not limited to consolidation, cavity, fibrosis, nodule, mass, calcification, interstitial thickening).''} 
While informative, such definitions contain extraneous details that do not emphasize the core visual characteristics of abnormalities. To address this, we define a set of visual attributes commonly used in medical imaging, including \{shape, location, density, color\}, which are crucial for characterizing an abnormality’s visual appearance. Note that although pixel intensity is a more accurate term in medical imaging, we retain color to ensure compatibility with generalist vision-language models that may not recognize the relationship between tissue density and intensity values. With both the definitions and visual attributes at hand, we design a prompting strategy to instruct the language model~\cite{achiam2023gpt} to generate decomposed knowledge descriptions focusing on the visual aspects of each abnormality. Specifically, we construct an input template that integrates medical definitions with defined visual features. The prompt follows this structure:  

\sloppy
\texttt{Here is the medical definition of [abnormality name]: "[medical-
definition]." Based on this definition and focusing on the follow- \\
ing visual attributes (e.g., shape, location, density, color), pro-\\
vide a brief description of the abnormality.}
\fussy

By prompting the large language model with this query, we obtain descriptions such as:  
\textit{``An area of increased density in the lung fields, typically appearing as a white or grayish patch.''} for each abnormality term.

\subsubsection{Architecture.} As shown in Figure \ref{method}B, we use the Florence-2 base \cite{xiao2024florence} as the backbone, which integrates a visual encoder and a multi-modal encoder-decoder. The visual encoder, based on DaViT\cite{ding2022davit}, processes an input image $I \in \mathbb{R}^{H \times W \times 3}$ to produce flattened visual token embeddings $V \in \mathbb{R}^{N \times D}$, where $N$ is the number of visual tokens and $D$ is their dimensionality. Simultaneously, knowledge-decomposed prompts are tokenized into text embeddings. Both the visual and text embeddings are passed through the multi-modal transformer encoder-decoder\cite{waswani2017attention} to generate the final answer.
The model generates output through auto-regressive decoding, applying cross-entropy loss to all discrete localization tokens. Finally, the loss is defined as:  
$$\mathcal{L} = - \sum_{i=1}^{N} \log p(y_i | \{V, T\})$$  
where $y_i$ is the target localization token at position $i$, and $p(y_i | \{V, T\})$ is the predicted probability distribution over the vocabulary, conditioned on both the visual and textual embedding. We fine-tune the entire model in an end-to-end manner using our decomposed textual knowledge prompts, which break down complex medical concepts into key visual attributes, guiding the model to focus on the core visual characteristics of abnormalities.

\section{Experiments}
\noindent \textbf{Dataset.}  
We trained our method on the VinDr-CXR dataset \cite{nguyen2022vindr}, a large-scale chest X-ray dataset with bounding boxes annotated by radiologists for various abnormalities. To ensure annotation consistency, we applied weighted box fusion \cite{muller2024chex} to merge overlapping bounding boxes and converted them into localization tokens. Since our task focuses on abnormality grounding, we retained only images with at least one annotated abnormality, resulting in 18,195 image-abnormality pairs, with 16,087 for training and 2,108 for test.
 To assess the zero-shot generalization capabilities of our method, we conducted experiments on the PadChest-GR dataset \cite{castro2024padchest}. We focused on two zero-shot scenarios: generalization to a new dataset and detection of previously unseen diseases. Following the predefined data split, we selected the test set and converted its bounding box annotations into text-box pairs, resulting in 1,285 image–bounding box pairs. To distinguish between these scenarios, we further divided the test set into two subsets. The first subset, PadChest-known (641 pairs), contains six diseases that overlap with the VinDr-CXR dataset. The second subset, PadChest-unknown (644 pairs), includes diseases not present in VinDr-CXR, enabling us to assess the model's performance in detecting previously unseen abnormalities. \\

\noindent \textbf{Benchmark Baselines.} We consider two recent state-of-the-art medical VLMs, MAIRA-2\cite{bannur2024maira} and RadVLM\cite{deperrois2025radvlm}, for comparison, both of which are significantly larger (with 30$\sim$56 times more parameters) than our model and trained on extensive multi-source datasets. Table \ref{comparision_results} shows that MAIRA-2 is a 13B parameter model trained on 501,825 training samples from a combination of MIMIC-CXR~\cite{johnson2019mimic}, PadChest~\cite{bustos2020padchest}, and USMix~\cite{demner2016preparing}. RadVLM, a 7B parameter model, is trained on an even broader set, including MIMIC-CXR, CheXpert-Plus~\cite{chambon2024chexpert}, CheXpert~\cite{irvin2019chexpert}, Vindr-CXR, MS-CXR~\cite{boecking2022making}, and PadChest-GR~\cite{castro2024padchest}, with a total of 1,022,742 image-instruction pairs. Our proposed model is trained solely on Vindr-CXR with 16,086 training samples, and has only 0.23B parameters.\\

\noindent \textbf{Experimental setup.} Our method leverages Florence-2-base~\cite{xiao2024florence} as the backbone, a 0.23B parameter architecture. We fine-tune the model using its pre-trained weights with the Adam optimizer~\cite{loshchilov2017decoupled}, setting the learning rate to $5 \times 10^{-6}$ and weight decay to 0.01. Training is conducted with a batch size of 16 and an input resolution of $512 \times 512$ on two NVIDIA A6000 GPUs. 
For comparison, we use the publicly available pre-trained weights of MAIRA-2 and RadVLM and evaluate them on the same test set.
In our ablation studies, we first establish a baseline model that uses only abnormality labels, without knowledge-enhanced prompts. The final model, in contrast, incorporates these prompts during training, which provide additional visual context and improve abnormality grounding.\\

\noindent \textbf{Evaluation metrics.} We evaluate all models using standard abnormality detection metrics~\cite{padilla2020survey}, including mean average precision (mAP) at various Intersection over Union (IoU) thresholds: mAP50, mAP50:95, and mAP75.
Besides, we use the Robust Detection Outcome (RoDeO)~\cite{meissen2023robust}, a metric for pathology detection that evaluates classification, shape, and localization for bounding box quality.

\section{Results}

\begin{table}[!t]

\centering
\setlength{\tabcolsep}{5pt}
\renewcommand{\arraystretch}{1.5}

\caption{Comparison on the VinDr-CXR and PadChest-GR datasets. Our method achieves competitive results on both datasets, with the best performance on VinDr-CXR and competitive results in a \textbf{zero-shot} setting on PadChest-GR. Best and second-best performances are coloured \colorbox{green!15}{Green} and \colorbox{yellow!15}{Yellow}. R$_{loc}$, R$_{shape}$, R$_{cls}$, and R$_{total}$ are different aspects of the RoDeO metrics. Methods marked with \includegraphics[width=0.025\linewidth]{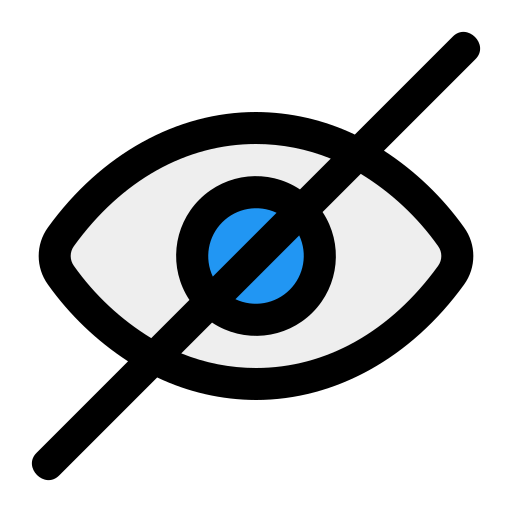} indicate the dataset had not been seen during training.}

\begin{adjustbox}{width=0.98\linewidth} 
\begin{tabular}{l | l | r | r | c c c | c c c c}
\toprule
Test Set & Method & \multicolumn{1}{c|}{Params} & \multicolumn{1}{c|}{Train. Samp.} & mAP$_{50}^{95}$ & mAP$_{50}$ & mAP$_{75}$ & R$_{loc}$ & R$_{shape}$ & R$_{cls}$ & R$_{total}$ \\
\midrule
\multirow{3}{*}{VinDr-CXR} 
&  MAIRA-2\includegraphics[width=0.03\linewidth]{images/eye.png}  & \cellcolor{pink!70}13B & 501,825& 1.22 & 4.94 & 0.32 & 25.65 & 17.23 & 80.13 & 24.08 \\
&  RadVLM & \cellcolor{pink!40}7B & 1,022,742 & \cellcolor{yellow!15}8.18 & \cellcolor{yellow!15}21.93 & \cellcolor{yellow!15}5.14 & \cellcolor{green!15}\textbf{60.15} & \cellcolor{yellow!15}39.47 & \cellcolor{green!15}\textbf{81.36} & \cellcolor{yellow!15}53.98 \\
&  Ours & \cellcolor{pink!10}0.23B & 16,087 & \cellcolor{green!15}\textbf{10.81} & \cellcolor{green!15}\textbf{25.5} & \cellcolor{green!15}\textbf{7.45} & \cellcolor{yellow!15}56.92 & \cellcolor{green!15}\textbf{41.41} & \cellcolor{yellow!15}80.92 & \cellcolor{green!15}\textbf{54.38} \\
\midrule
\multirow{3}{*}{PadCh. Know.} 
& MAIRA-2  & \cellcolor{pink!70}13B & 501,825 & \cellcolor{green!15}\textbf{8.36} & \cellcolor{green!15}\textbf{19.17} & \cellcolor{yellow!20}\textbf{5.81} & 33.05 & \cellcolor{yellow!15}29.68 & \cellcolor{yellow!15}81.92 & 37.14 \\
& RadVLM  &  \cellcolor{pink!40}7B& 1,022,742  & 2.53 & 10.84 & \cellcolor{green!15}0.81 & \cellcolor{green!15}\textbf{58.61} & 29.18 & 79.64 & \cellcolor{yellow!15}46.16 \\
& Ours\includegraphics[width=0.033\linewidth]{images/eye.png}  &  \cellcolor{pink!10}0.23B & 16,087 & \cellcolor{yellow!15}2.68 & \cellcolor{yellow!15}11.07 & 0.56 & \cellcolor{yellow!15}57.34 & \cellcolor{green!15}\textbf{32.48} & \cellcolor{green!15}\textbf{83.00} & \cellcolor{green!15}\textbf{49.13} \\
\bottomrule
\end{tabular}
\label{comparision_results}
\end{adjustbox}
\end{table}

\begin{figure*}[!t]
\centerline{\includegraphics[width=\textwidth]{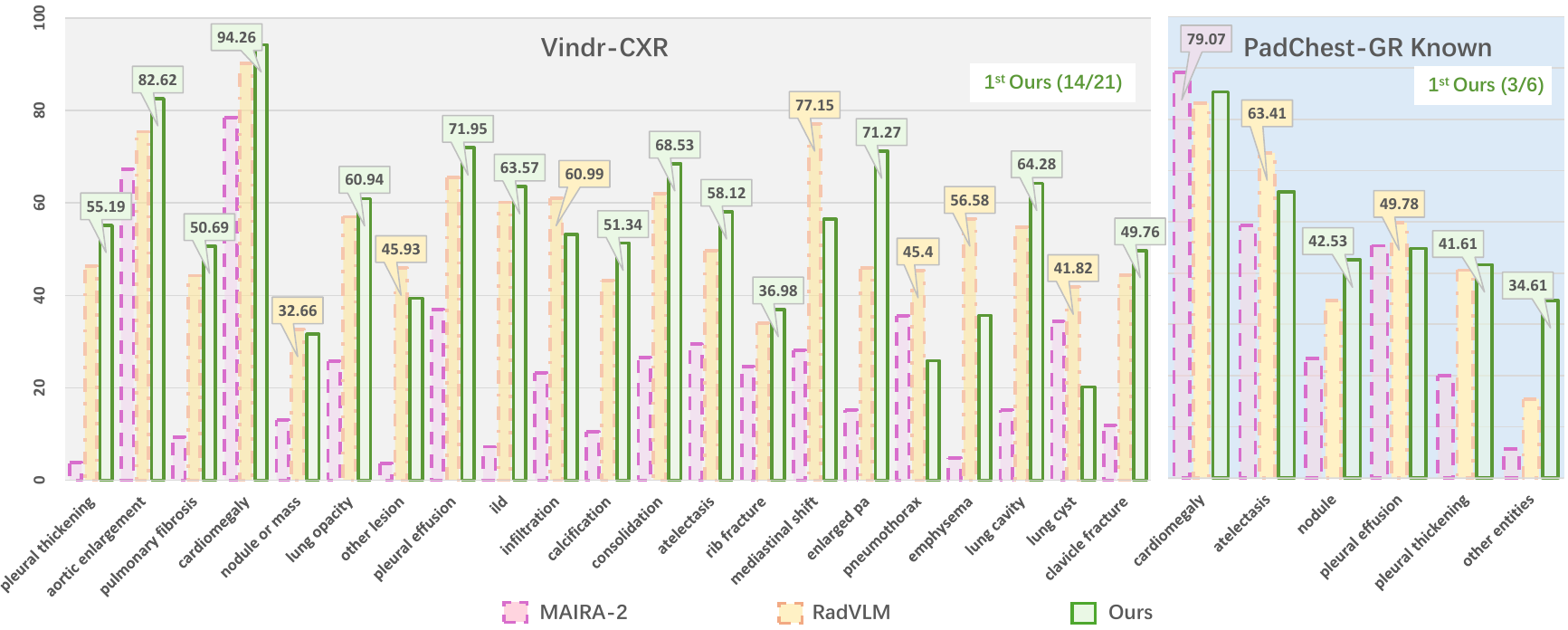}}
\caption{Performance for each disease class, with the y-axis representing the RoDeo total metric. Our method achieves first place in 14 out of 21 diseases from the VinDr-CXR dataset and 3 out of 6 known diseases from the PadChest-GR dataset. The best performances are highlighted in the callout.}
\label{fig1}
\label{each_disease}
\end{figure*}

\subsection{Comparison with SOTA}
\noindent \textbf{Fewer Parameters, Competitive Performance.} 
Table \ref{comparision_results} compares the performance of our method with existing state-of-the-art models on the VinDr-CXR and PadChest-Known datasets. Despite having only 0.23B parameters, our model consistently outperforms much larger counterparts across multiple key evaluation metrics on the VinDr-CXR dataset. Specifically, our method achieves the highest mAP50 of 25.5\%, surpassing the second-best model by 3.57 points. It also attains a mAP75 of 7.45\%, outperforming the second-best model by 2.31 points. For mAP50:95, our model reaches 10.81\%, exceeding the closest competitor by 2.63 points. Additionally, our method achieves the highest overall RoDeO score of 54.38\%. These results demonstrate that, despite having significantly fewer parameters, our knowledge-enhanced model achieves competitive, and in some cases, superior performance compared to much larger models trained on extensive multi-source datasets. Notably, we observe that MAIRA-2 underperforms significantly compared to RadVLM, which can be attributed to MAIRA-2 not being trained on the VinDr-CXR dataset, thus operating in a zero-shot setting. 
In Figure \ref{each_disease}, we also evaluate our method's performance across individual diseases. Our model ranks first in 14 out of 21 diseases on VinDr-CXR, highlighting its ability to effectively detect abnormalities across a wide range of conditions. \\

\noindent \textbf{Comparable Zero-shot Performance.} We evaluate our method on the Pad-Chest-Known dataset, which contains 6 diseases overlapping with the Vindr-CXR dataset, in a zero-shot setting. Table \ref{comparision_results} shows that our method achieves the best performance in RoDeO shape matching, classification, and overall scores, with values of 32.48\%, 83.00\%, and 49.13\%, respectively. Besides, our model ranks second in mAP50:95 with a score of 2.68\% and in mAP50 with 11.07\%. Our method, though not trained on PadChest, shows competitive performance against MAIRA-2 and RadVLM.
We also evaluate the performance for each disease, where our model ranks first in 3 out of 6 classes, demonstrating its good generalization capability, as shown in the right part of the Figure \ref{each_disease}.

\subsection{Ablation Study}
\noindent \textbf{Knowledge-enhanced prompts boost VLM's performance.}
Table \ref{ablation_table} pre-sents the results of an ablation study comparing the baseline to the knowledge-enhanced method on the Vindr-CXR and PadChest-GR datasets. Our proposed method demonstrates substantial improvements across all evaluation metrics on the Vindr-CXR test set, achieving a mAP50 of 25.5\% (vs. 13.26\%), mAP75 of 7.45\% (vs. 2.76\%), and an overall RoDeO score of 54.38\% (vs. 45.22\%). These results highlight that integrating disease-specific visual knowledge enhances the model's ability to detect abnormalities effectively, outpacing the baseline by a large margin. We also evaluate both methods' performance on the PadChest-GR Known dataset. While mAP75 shows a slight decrease compared to the baseline, all other metrics exhibit considerable improvement, indicating that the knowledge integration allows the model to generalize better to different datasets. \\

\noindent \textbf{Knowledge-enhanced prompts improve detection of unknown findings.} In this section, we further evaluate the performance of our proposed method on the PadChest-Unknown dataset, which contains 18 diseases not present in the Vindr-CXR dataset. Table \ref{ablation_table} shows that the knowledge-enhanced method outperforms the baseline across all metrics. Specifically, mAP50 increases from 1.48\% to 3.05\%, and mAP75 improves from 0.03\% to 0.29\%. These results suggest that integrating disease-specific knowledge enhances the model's ability to transfer knowledge to unknown diseases, i.e., not encountered during training, thereby improving zero-shot performance.

\begin{table}[!t]
\centering

\setlength{\tabcolsep}{6pt}
\renewcommand{\arraystretch}{1.0}
\caption{Ablation study on the effect of knowledge descriptions. \textit{Base} refers to the Florence-2 model~\cite{xiao2024florence}, while \textit{Ours} incorporates knowledge descriptions (KD). We evaluate in-distribution performance on VinDr-CXR and assess zero-shot generalization to an unseen dataset (PadChest-Known, marked by \includegraphics[width=0.023\linewidth]{images/eye.png}) and to both an unseen dataset and previously unseen disease classes (PadChest-Unknown, marked by \includegraphics[width=0.023\linewidth]{images/eye.png}\includegraphics[width=0.023\linewidth]{images/eye.png}). Best performances are highlighted in \colorbox{green!15}{green}. }

\begin{adjustbox}{width=1.00\linewidth} 
\begin{tabular}{l | l | c c c | c c c c}
\toprule
Test Set & Method & mAP$_{50}^{95}$ & mAP$_{50}$ & mAP$_{75}$ & R$_{loc}$ & R$_{shape}$ & R$_{cls}$ & R$_{total}$ \\
\midrule
\multirow{2}{*}{Vindr-CXR} & Base & 4.92 & 13.26 & 2.76 & 44.26 & 34.36 & 78.19 & 45.22 \\
 & + KD (Ours) &  \cellcolor{green!15}\textbf{10.81} &  \cellcolor{green!15}\textbf{25.50} &  \cellcolor{green!15}\textbf{7.45} &  \cellcolor{green!15}\textbf{56.92} &  \cellcolor{green!15}\textbf{41.41} &  \cellcolor{green!15}\textbf{80.92} &  \cellcolor{green!15}\textbf{54.38} \\
\midrule
\midrule
\multirow{1}{*}{Pad. Know.} & Base & 1.92 & 8.34 &  \cellcolor{green!15}\textbf{0.65} & 48.11 & 31.19 & 81.25 & 44.59 \\
\includegraphics[width=0.023\linewidth]{images/eye.png} & + KD (Ours)  &  \cellcolor{green!15}\textbf{2.68} &  \cellcolor{green!15}\textbf{11.07} & 0.56 &  \cellcolor{green!15}\textbf{57.34} &  \cellcolor{green!15}\textbf{32.48} &  \cellcolor{green!15}\textbf{83.00} &  \cellcolor{green!15}\textbf{49.13} \\
\midrule
\multirow{1}{*}{Pad. Unkn.} & Base & 0.37 & 1.48 & 0.03 & 38.14 & 20.90 & 78.69 & 32.05 \\
\includegraphics[width=0.023\linewidth]{images/eye.png}\includegraphics[width=0.023\linewidth]{images/eye.png} & + KD (Ours) &  \cellcolor{green!15}\textbf{0.95} &  \cellcolor{green!15}\textbf{3.05} &  \cellcolor{green!15}\textbf{0.29} &  \cellcolor{green!15}\textbf{44.71} &  \cellcolor{green!15}\textbf{22.93} &  \cellcolor{green!15}\textbf{86.12} &  \cellcolor{green!15}\textbf{33.72} \\
\bottomrule
\end{tabular}
\end{adjustbox}
\label{ablation_table}
\end{table}

\section{Discussion and Conclusion}
We introduce a novel knowledge-enhanced approach to VLMs for abnormality grounding. By integrating fine-grained, decomposed disease-specific visual descriptions, our method demonstrates that smaller, task-specific models can achieve performance comparable to much larger VLMs trained on extensive datasets. This approach substantially improves zero-shot generalization, particularly for previously unseen datasets, making VLMs more effective in low-data settings. Our results further demonstrate that knowledge-enhanced prompts not only boost in-domain performance but also help the model generalize to new diseases, highlighting their potential for real-world medical applications where labeled data is scarce.

While our results demonstrate the effectiveness of our knowledge-enhanced prompts method, several avenues for future research remain. First, expanding the knowledge base to include a broader range of diseases, along with the integration of multimodal data sources, could further enhance the model's generalization capability. 
Additionally, integrating decomposed, knowledge-enhanced prompts with larger, more complex VLMs could push performance boundaries. Larger models trained on more extensive datasets have the potential to benefit from the specialized disease-specific knowledge embedded in the prompts, improving their performance.
Finally, exploring dynamic prompt adjustment for each disease could further optimize model performance. Recent studies have shown that prompt engineering, which focuses on emphasizing the most relevant cues for each disease, can significantly enhance VLMs performance.

\bibliographystyle{splncs04}
\bibliography{MyLibrary}

\begin{thebibliography}{10}
\providecommand{\url}[1]{\texttt{#1}}
\providecommand{\urlprefix}{URL }
\providecommand{\doi}[1]{https://doi.org/#1}

\bibitem{openai2024chatgpt}
OpenAI: {ChatGPT} can now see, hear, and speak. \url{https://openai.com/index/chatgpt-can-now-see-hear-and-speak/} (2024), accessed: 2024-11-26

\bibitem{xiao2024florence}
Xiao, B., Wu, H., Xu, W., et~al.: Florence-2: Advancing a unified representation for a variety of vision tasks. In: Proceedings of the IEEE/CVF Conference on Computer Vision and Pattern Recognition. pp. 4818--4829 (2024)

\bibitem{bai2023qwen}
Bai, J., Bai, S., Yang, S., Wang, S., Tan, S., Wang, P., Lin, J., Zhou, C., Zhou, J.: Qwen-vl: A frontier large vision-language model with versatile abilities. arXiv preprint arXiv:2308.12966  (2023)

\bibitem{lu2024deepseek}
Lu, H., Liu, W., Zhang, B., Wang, B., Dong, K., Liu, B., Sun, J., Ren, T., Li, Z., Yang, H., et~al.: {DeepSeek-vl}: towards real-world vision-language understanding. arXiv preprint arXiv:2403.05525  (2024)

\bibitem{stefanini2022show}
Stefanini, M., Cornia, M., et~al.: From show to tell: A survey on deep learning-based image captioning. IEEE transactions on pattern analysis and machine intelligence  \textbf{45}(1),  539--559 (2022)

\bibitem{lin2023medical}
Lin, Z., Zhang, D., et~al.: Medical visual question answering: A survey. Artificial Intelligence in Medicine  \textbf{143},  102611 (2023)

\bibitem{xiao2024towards}
Xiao, L., Yang, X., Lan, X., et~al.: Towards visual grounding: A survey. arXiv preprint arXiv:2412.20206  (2024)

\bibitem{deperrois2025radvlm}
Deperrois, N., Matsuo, H., Ruip{\'e}rez-Campillo, S., Vandenhirtz, M., Laguna, S., Ryser, A., Fujimoto, K., Nishio, M., Sutter, T.M., Vogt, J.E., et~al.: {RadVLM}: A multitask conversational vision-language model for radiology. arXiv preprint arXiv:2502.03333  (2025)

\bibitem{bannur2024maira}
Bannur, S., Bouzid, K., Castro, D.C., Schwaighofer, A., Thieme, A., Bond-Taylor, S., Ilse, M., P{\'e}rez-Garc{\'\i}a, F., Salvatelli, V., Sharma, H., et~al.: Maira-2: Grounded radiology report generation. arXiv preprint arXiv:2406.04449  (2024)

\bibitem{chen2024chexagent}
Chen, Z., Varma, et~al.: Chexagent: Towards a foundation model for chest {X-ray} interpretation. arXiv preprint arXiv:2401.12208  (2024)

\bibitem{johnson2019mimic}
Johnson, A.E., Pollard, T.J., Greenbaum, N.R., Lungren, M.P., Deng, C.y., Peng, Y., Lu, Z., Mark, R.G., Berkowitz, S.J., Horng, S.: {MIMIC-CXR-JPG}, a large publicly available database of labeled chest radiographs. arXiv preprint arXiv:1901.07042  (2019)

\bibitem{wu2021chest}
Wu, J.T., Agu, N.N., Lourentzou, I., Sharma, A., Paguio, J.A., Yao, J.S., Dee, E.C., Mitchell, W., Kashyap, S., Giovannini, A., et~al.: Chest imagenome dataset for clinical reasoning. arXiv preprint arXiv:2108.00316  (2021)

\bibitem{bustos2020padchest}
Bustos, A., Pertusa, A., Salinas, J.M., De~La Iglesia-Vaya, M.: Padchest: A large chest x-ray image dataset with multi-label annotated reports. Medical Image Analysis  \textbf{66},  101797 (2020)

\bibitem{castro2024padchest}
Castro, D.C., Bustos, A., Bannur, S., Hyland, S.L., Bouzid, K., Wetscherek, M.T., S{\'a}nchez-Valverde, M.D., Jaques-P{\'e}rez, L., P{\'e}rez-Rodr{\'\i}guez, L., Takeda, K., et~al.: Padchest-gr: A bilingual chest {X-ray} dataset for grounded radiology report generation. arXiv preprint arXiv:2411.05085  (2024)

\bibitem{li2024ultrasound}
Li, J., Su, T., Zhao, B., Lv, F., Wang, Q., Navab, N., Hu, Y., Jiang, Z.: Ultrasound report generation with cross-modality feature alignment via unsupervised guidance. arXiv preprint arXiv:2406.00644  (2024)

\bibitem{tanida2023interactive}
Tanida, T., M{\"u}ller, P., Kaissis, G., Rueckert, D.: Interactive and explainable region-guided radiology report generation. In: Proceedings of the IEEE/CVF Conference on Computer Vision and Pattern Recognition. pp. 7433--7442 (2023)

\bibitem{li2022self}
Li, J., Li, S., Hu, Y., Tao, H.: A self-guided framework for radiology report generation. In: International Conference on Medical Image Computing and Computer-Assisted Intervention. pp. 588--598. Springer (2022)

\bibitem{singhal2025toward}
Singhal, K., Tu, T., Gottweis, J., Sayres, R., Wulczyn, E., Amin, M., Hou, L., Clark, K., Pfohl, S.R., Cole-Lewis, H., et~al.: Toward expert-level medical question answering with large language models. Nature Medicine pp.~1--8 (2025)

\bibitem{li2025language}
Li, J., Kim, S.H., M{\"u}ller, P., Felsner, L., Rueckert, D., Wiestler, B., Schnabel, J.A., Bercea, C.I.: Language models meet anomaly detection for better interpretability and generalizability. In: Medical Image Computing and Computer Assisted Intervention–MICCAI 2024 Workshops. Lecture Notes in Computer Science, vol. 15401, pp. 1--11. Springer Nature Switzerland AG (2025)

\bibitem{zhang2023pmc}
Zhang, X., Wu, C., Zhao, Z., et~al.: {Pmc-VQA}: Visual instruction tuning for medical visual question answering. arXiv preprint arXiv:2305.10415  (2023)

\bibitem{jiang2022review}
Jiang, P., Ergu, D., et~al.: A review of yolo algorithm developments. Procedia computer science  \textbf{199},  1066--1073 (2022)

\bibitem{nguyen2022vindr}
Nguyen, H.Q., Lam, K., Le, L.T., Pham, H.H., Tran, D.Q., Nguyen, D.B., Le, D.D., Pham, C.M., Tong, H.T., Dinh, D.H., et~al.: Vindr-cxr: An open dataset of chest {X-rays} with radiologist’s annotations. Scientific Data  \textbf{9}(1), ~429 (2022)

\bibitem{achiam2023gpt}
Achiam, J., Adler, S., Agarwal, S., Ahmad, L., Akkaya, I., Aleman, F.L., Almeida, D., Altenschmidt, J., Altman, S., Anadkat, S., et~al.: {GPT-4} technical report. arXiv preprint arXiv:2303.08774  (2023)

\bibitem{ding2022davit}
Ding, M., Xiao, B., Codella, N., et~al.: Davit: Dual attention vision transformers. In: European Conference on Computer Vision. pp. 74--92. Springer (2022)

\bibitem{waswani2017attention}
Waswani, A., Shazeer, N., Parmar, N., Uszkoreit, J., Jones, L., Gomez, A., Kaiser, L., Polosukhin, I.: Attention is all you need. In: NIPS (2017)

\bibitem{muller2024chex}
M{\"u}ller, P., Kaissis, G., Rueckert, D.: Chex: Interactive localization and region description in chest {X-ray}. In: European Conference on Computer Vision. pp. 92--111. Springer (2024)

\bibitem{demner2016preparing}
Demner-Fushman, D., Kohli, M.D., Rosenman, M.B., Shooshan, S.E., Rodriguez, L., Antani, S., Thoma, G.R., McDonald, C.J.: Preparing a collection of radiology examinations for distribution and retrieval. Journal of the American Medical Informatics Association  \textbf{23}(2),  304--310 (2016)

\bibitem{chambon2024chexpert}
Chambon, P., Delbrouck, J.B., Sounack, T., Huang, S.C., Chen, Z., Varma, M., Truong, S.Q., Chuong, C.T., Langlotz, C.P.: {CheXpert Plus}: Augmenting a large chest {X-ray} dataset with text radiology reports, patient demographics and additional image formats. arXiv preprint arXiv:2405.19538  (2024)

\bibitem{irvin2019chexpert}
Irvin, J., Rajpurkar, P., Ko, M., Yu, Y., Ciurea-Ilcus, S., Chute, C., Marklund, H., Haghgoo, B., Ball, R., Shpanskaya, K., et~al.: Chexpert: A large chest radiograph dataset with uncertainty labels and expert comparison. In: Proceedings of the AAAI conference on artificial intelligence. vol.~33, pp. 590--597 (2019)

\bibitem{boecking2022making}
Boecking, B., Usuyama, N., Bannur, S., Castro, D.C., Schwaighofer, A., Hyland, S., Wetscherek, M., Naumann, T., Nori, A., Alvarez-Valle, J., et~al.: Making the most of text semantics to improve biomedical vision--language processing. In: European Conference on Computer Vision. pp. 1--21. Springer (2022)

\bibitem{loshchilov2017decoupled}
Loshchilov, I., Hutter, F.: Decoupled weight decay regularization. In: International Conference on Learning Representations (2018)

\bibitem{padilla2020survey}
Padilla, R., Netto, S.L., Da~Silva, E.A.: A survey on performance metrics for object-detection algorithms. In: 2020 International Conference on Systems, Signals and Image Processing (IWSSIP). pp. 237--242. IEEE (2020)

\bibitem{meissen2023robust}
Meissen, F., M{\"u}ller, P., Kaissis, G., et~al.: Robust detection outcome: A metric for pathology detection in medical images. In: Medical Imaging with Deep Learning

\end{thebibliography}
\end{document}